\RequirePackage{tikz}
\documentclass[default]{sn-jnl}

\usepackage{xcolor}

\definecolor{AUC}{HTML}{D7191C}
\definecolor{MRR}{HTML}{FDAE61}
\definecolor{NDCG@5}{HTML}{ABDDA4}
\definecolor{NDCG@10}{HTML}{2B83BA}
\usepackage{array,multirow}
\usepackage{graphicx}
\usepackage[T1]{fontenc}
\usepackage{multirow}
\usepackage{subfigure}
\usepackage{adjustbox}
\usepackage{pgfplots}
\usepackage{graphicx}
\usepackage[labelsep=endash]{caption}
\usepackage{amssymb}
\usepackage{float}
\usepackage{graphicx,stackengine,scalerel}

\usepackage{amssymb}
\usepackage{mathtools}
\usepackage{bm}

\jyear{2023}
\begin{document}
\title{DOR: A Novel Dual-Observation-Based Approach for News Recommendation Systems}
	%
	%
	%
	%
	%

	\author*[1]{\fnm{Mengyan} \sur{Wang}}\email{mengyan.wang@autuni.ac.nz}
	
	\author*[1]{\fnm{Weihua} \sur{Li}}\email{weihua.li@aut.ac.nz}
	
	\author[1]{\fnm{Jingli} \sur{Shi}}\email{Jingli.shi@aut.ac.nz}
	
	\author[2]{\fnm{Shiqing} \sur{Wu}}\email{shiqing.wu@uts.edu.au}
	
	\author[3]{\fnm{Quan} \sur{Bai}}\email{quan.bai@utas.edu.au}

	\affil*[1]{\orgdiv{School of Engineering, Computer \& Mathematical Sciences}, \orgname{Auckland University of Technology}, \orgaddress{\street{55 Wellesley Street East}, \city{Auckland}, \postcode{1010}, \country{New Zealand}}}
	
	\affil[2]{\orgdiv{School of Computer Science}, \orgname{University of Technology Sydney}, \orgaddress{\street{15 Broadway Rd, Ultimo}, \city{Sydney}, \postcode{2007}, \state{NSW}, \country{Australia}}}
	
	\affil[3]{\orgdiv{School of Technology, Environments and Design}, \orgname{University of Tasmania}, \orgaddress{\street{Churchill Ave}, \city{Hobart}, \postcode{7005}, \state{Tasmania}, \country{Australia}}}

\abstract{
Online social media platforms offer access to a vast amount of information, but sifting through the abundance of news can be overwhelming and tiring for readers. personalised recommendation algorithms can help users find information that interests them. However, most existing models rely solely on observations of user behaviour, such as viewing history, ignoring the connections between the news and a user's prior knowledge. This can result in a lack of diverse recommendations for individuals. In this paper, we propose a novel method to address the complex problem of news recommendation. Our approach is based on the idea of dual observation, which involves using a deep neural network with observation mechanisms to identify the main focus of a news article as well as the focus of the user on the article. This is achieved by taking into account the user's belief network, which reflects their personal interests and biases. By considering both the content of the news and the user's perspective, our approach is able to provide more personalised and accurate recommendations. We evaluate the performance of our model on real-world datasets and show that our proposed method outperforms several popular baselines.
}


\maketitle

\section{Introduction}
    
    In recent years, the proliferation of online news websites has made it easy for people to access global news, but the sheer amount of information available on social media can make it difficult for users to find articles that are of interest to them. To address this problem, news recommendation systems have become popular \cite{wu2019npa}.

    Over the past two decades, many researchers have focused on developing recommendation systems that create personalised reading lists based on users' reading behaviour. Traditional methods such as content-based filtering (CBF), collaborative filtering (CF), and hybrid recommendation systems have been successful in many cases \cite{liu2010personalized}. However, these methods can be limited by data sparsity, and the cold start problem \cite{wang2018ripplenet}. To overcome these challenges, some researchers have enriched models with additional information and used deep neural networks to model users' interests more accurately. Nevertheless, these approaches may not fully capture low-order propagation information \cite{wang2018ripplenet}. In order to capture contextual information from users' reading behaviour and obtain more users' interests, low-order propagation information should be incorporated into the research. The low-order propagation information indicates enriched features of each user's behaviour, which is obtained by deep neural networks \cite{lei2020hybrid}.
    
    Recently, there has been a trend towards combining knowledge graphs with deep neural networks to consider low and high-order relations in news recommendation simultaneously \cite{wang2018dkn,wang2019knowledge}. For example, Wang et al. adopted deep learning networks (TransE) to capture low-order relations representations and employed graph neural networks to capture high-order relations representations \cite{wang2019knowledge}. Despite these advances, the recommendation task remains challenging for a number of reasons as below. 

	\begin{itemize}
		
		\item Different readers may focus on different aspects of the same article for different reasons. For example, a news headline such as \textit{"New Zealand fully reopens to the world in August: Ardern"} may attract readers who are interested in the economic impacts of the border reopening, as well as those who are interested in the real estate market. Therefore, it is important to consider the various belief networks of readers in the recommendation process.

        \item One-sided or incomplete observations may not be sufficient to form a satisfactory user belief system. A user's belief system is shaped by all the information they have encountered. This information should include all relevant contextual semantics and should be revised as the user's beliefs change. Therefore, it is important to leverage dual observation mechanisms to deeply explore the semantic information in the news and analyse the influences of each news article on the user's belief network.
		
    \end{itemize}
	
	To address the two challenging issues of news recommendation mentioned above, this paper proposes a novel approach called the dual-observation-based approach for news recommendation (DOR). The DOR architecture incorporates an internal and external observation-based interest extraction model, which learns and extracts both internal and external users' interest expressions. By integrating dual observation networks, the DOR approach can endow each piece of user behaviour with deeper meaning. Specifically, the internal observation network is a content-based learning model that imports news titles and content, forming the user's belief networks. The external observation network, on the other hand, explores the mutual influence between the users' belief networks and various extrinsic information sources. By considering both internal and external observations, the DOR approach is able to provide more accurate and personalised recommendations to users.


This research paper makes a couple of contributions. Firstly, it proposes a novel dual-observation networks-based recommendation architecture. Our proposed model focuses on exploring more semantic information from a piece of news and constructing more satisfying user preferences by employing specific news expressions and users' belief networks. Secondly, the complex combination of low and high-order relation expressions further avoids the limitations of data leakage and the cold-start problem.

 The remainder of this paper is organised as follows: Section 2 reviews literature related to this study. In Section 3, we introduce some preliminary concepts of our research and the problem description. Subsequently, the details of our proposed dual-observation-based recommendation system are given in Section 4. Section 5 describes the experimental settings and demonstrates the results of the experiments. At last, we conclude this research work and point out the future direction.

\section{Related Works}

\subsection{Feature-based Recommendation}

Recommendation systems have been a popular research topic for many years, and there have traditionally been two main techniques for studying them: feature-based techniques and deep learning-based techniques \cite{wang2017dynamic}. One well-known feature-based modelling technique is collaborative filtering (CF) \cite{su2009survey}, which has been widely used to develop recommendation systems because it can effectively capture the interactions between users and items \cite{liang2008collaborative,rendle2012bpr}. However, CF-based algorithms may suffer from the cold-start problem and the issue of information sparsity, and may not provide satisfactory performance \cite{wang2019kgat}.

To address these limitations, content-based (CB) modelling algorithms were proposed \cite{pazzani2007content}. These algorithms calculate the similarity of content features and recommend similar content. Capelle et al. used a CB modelling technique to model items and employed the WordNet synonym set to replace TF-IDF for estimating semantic importance \cite{capelle2012semantics}. However, CB modelling algorithms are handcrafted and require extensive domain knowledge, which can be time-consuming \cite{wu2022personalized}. Lin et al. and Burke et al. proposed hybrid recommendation modelling algorithms that combine CF and CB algorithms \cite{lin2002efficient,burke2002hybrid}. However, these approaches do not completely solve the previous problems.

\subsection{Deep Learning-based Recommendation}

Deep learning has been widely adopted and applied to a variety of applications, including recommender systems \cite{aggarwal2018neural,park2017deep,zheng2018drn}. Zhang et al. summarised several classical deep neural networks for recommendations, including Multilayer Perceptron (MLP), Autoencoder (AE), Convolutional Neural Network (CNN), Recurrent Neural Network (RNN), Restricted Boltzmann Machine (RBM), Neural Autoregressive Distribution Estimation (NADE), Adversarial Network (AN), Attentional Models (AM), and Deep Reinforcement Learning (DRL) \cite{zhang2019deep}. These approaches offer several benefits for recommendation systems, such as the ability to model complex user-item interactions through nonlinear transformation (NT), the ability to learn rich item representations through representation learning (RL), the ability to model sequential user behaviour through sequence modelling (SM), and increased flexibility.

However, recommendation systems based solely on deep neural networks may not be able to explain complex interaction patterns fully and may be perceived as a "black box" due to their lack of interpretability \cite{zhang2019deep}. Okura et al. used an embedding-based model to represent items \cite{okura2017embedding}, while Park et al. proposed an RNN method to mimic user preferences \cite{park2017deep}. Zhu et al. developed a news recommendation system based on the RNN model called the Deep Attention Network (DAN) model, which demonstrated the importance of the RNN method in fully exploring users' historical sequential features \cite{zhu2019dan}. Chen et al. designed a Co-occurrence CNN that considers both user-item and item-item interactions \cite{chen2022cocnn}. Guo et al. used deep learning techniques to address the limitations of previous social recommendation research, such as insufficient robust data management and overly specific preferences \cite{guo2021deep}. These deep learning-based approaches can automatically discover information-rich item expressions without the need for extensive manual processes and often provide a better understanding of item content than feature-based techniques.

\subsection{Knowledge Graph-based Recommendation}

The integration of knowledge graphs into recommendation systems has received significant attention from researchers, resulting in several successful approaches. For example, Sun et al. demonstrated the effectiveness of knowledge graphs in improving recommendation satisfaction by using both knowledge graphs and DNNs to obtain item representations \cite{sun2018recurrent}. Wang et al. proposed a knowledge graph-based recommendation model that learned movie representations at a high-order level, achieving satisfactory results \cite{wang2019multi}. Jiang et al. addressed the cross-domain problem by constructing a social-domain graph and incorporating information from other domains into the social graph. They proposed a Hybrid Random Walk (HRW) method for social recommendations \cite{jiang2015social}. Fan et al. also integrated knowledge graphs into the recommendation model, using Graph Neural Networks (GNNs) to learn features from duplicate user-user and user-item graphs \cite{fan2019graph}.

In text-based scenarios, knowledge graphs have been used to extract semantic meaning. For example, Sheu et al. applied a knowledge graph to the news domain, focusing on exploring the contextual features of news to represent users' reading interests over a short time period. They used Graph Convolutional Networks (GCNs) to embed contextual information \cite{sheu2020context}. Wang et al. incorporated a knowledge graph into a news recommendation system for news content engineering, using TransE to represent news entities and pre-trained Word2Vec to express word embeddings. The results showed that the knowledge graph had a significant impact on the recommendation ability \cite{wang2018dkn}.

\subsection{Attention-based Recommendation Systems using Deep Neural Networks}

Recently, researchers have incorporated attention mechanisms into recommendation systems to improve performance. Attention is a method that helps a model identify which parts of the input data are most important for making a decision \cite{jiang2022san}. The attention method does the task of distinguishing the importance of data by learning a pattern in the data and using that pattern to focus on certain parts of the data more heavily when making a decision \cite{hekmatfar2022attention}. Attention mechanisms enable personalised recommendations by allowing the model to focus on relevant information and automatically extract relevant information, improving the understanding of the item's content \cite{vijaikumar2022neural}. 

Jung et al. used an In-and-Out Attention flow framework (AttnIO) in a dialogue recommendation system \cite{jung2020attnio}, while Zhu et al. developed an attention-based DNN news recommendation system \cite{zhu2019dan}. Wu et al. considered diverse news information in their recommendation model and included an attention mechanism \cite{wu2019neural}. They represented users' interests from word-level expressions and simultaneously incorporated category embeddings into the news embeddings. Li et al. also adopted a similar method, using an attention-based deep neural network recommendation system in various scenarios \cite{li2021deep}. Wang et al. leveraged the CNN model to perform relation classification for the recommendation task \cite{wang2016relation}, highlighting the non-trivial influence of relations in contextual representation learning. These studies demonstrate the positive impact of attention networks on recommendation research.

However, prior studies have not considered the interaction between the user's knowledge system and input information from a macro perspective and the context-rich semantic expression of input information.

\subsection{Summary}
Overall, existing recommendation systems may have difficulties adequately capturing the two-way influence between users' belief networks and external information sources and may have a limited understanding of these external sources, such as insufficient contextual representations. Furthermore, these systems can be prone to the cold-start problem and may lack transparency due to their black-box nature. Our proposed DOR model addresses these issues by constructing users' preferences using a dual-observation mechanism. The external observation examines the mutual influence between users' belief networks and various external information sources, while the internal observation extracts rich contextual semantic information. In addition, the DOR model learns user interests from both low-order and high-order mechanisms based on dual observation mechanisms, including internal and external agencies. The high-order expression model effectively avoids the limitation of cold-start and increases transparency in the recommendation process.

	

\section{Preliminary}

This section will introduce several preliminary concepts necessary for this research, including knowledge graph embedding and dual observations. Afterwards, we will formulate the problem in the current setting.

\subsection{Knowledge Graph Embedding}
Knowledge graphs have been widely studied from many perspectives, including representation and modelling, knowledge identification, knowledge fusion, and knowledge retrieval and reasoning \cite{chen2020review}. Benefiting from the power of knowledge graphs, incorporating knowledge graphs into recommendation systems has become popular in recent years, as it can significantly improve the performance of recommendations \cite{wang2017knowledge}. In general, a knowledge graph consists of a number of Resource Description Framework (RDF) triples, where each RDF triple contains a head entity $h$, a relation $r$, and a tail entity $t$ \cite{omran2019learning}. To effectively derive and utilise information on entities and relations in the knowledge graph, it is necessary to represent them as low-dimensional vectors in a continuous space, i.e., embeddings. These embeddings can be used for subsequent tasks, such as link prediction, entity classification, and knowledge base completion \cite{bordes2013translating}. 

There are several approaches for learning knowledge graph embeddings, such as neural network-based models \cite{wang2019knowledge}, semantic matching models \cite{bordes2013translating}, and translation-based models \cite{ji2015knowledge}. The translation-based models have proved their effectiveness and efficiency in representing entities and relations, and they have the complexity of space and time that scales linearly with the dimensionality of entities and relations embedding space \cite{wang2017knowledge}. Furthermore, neural network-based embedding models and semantic matching models usually suffer from the limitation of data sparsity and over-simplify \cite{mohamed2021biological}. Hence, we select three widely used translation-based models (i.e., TransE, TransH, and TransR) to represent knowledge graph triples into low-dimensional embeddings in the proposed DOR system. The details of these three models are as follows:

\begin{figure*}[t]
    \centering
    \subfigure[TransE]{\includegraphics[width=0.35\textwidth]{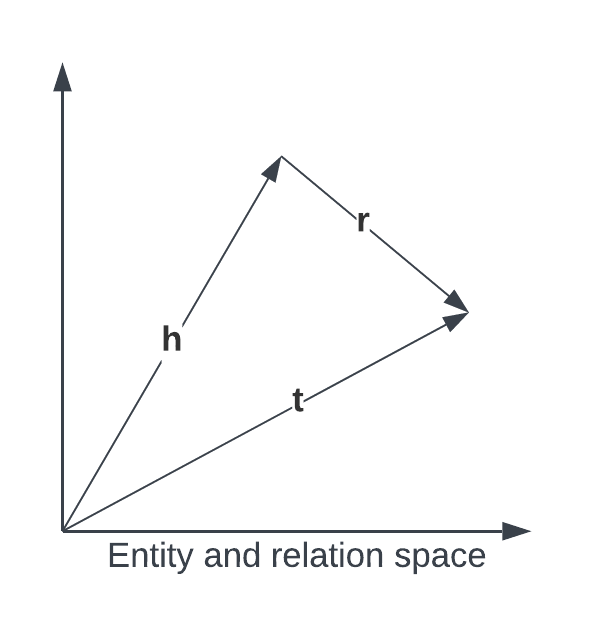}\label{fig:transE}}
    \subfigure[TransH]{\includegraphics[width=0.35\textwidth]{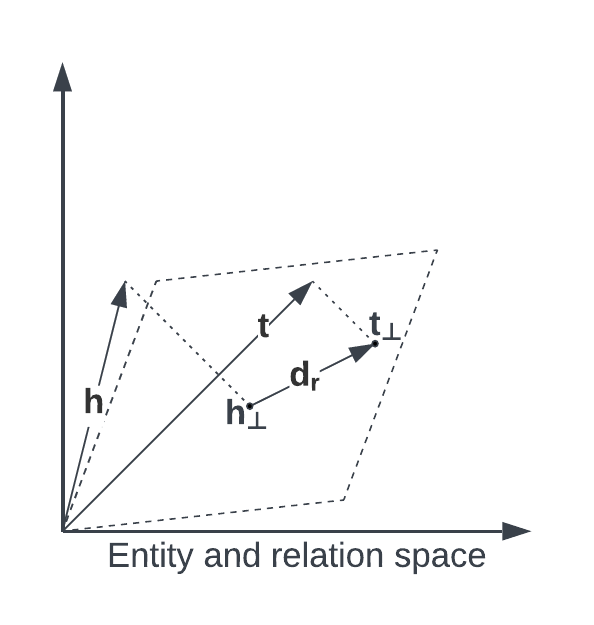}\label{fig:transH}}
    \subfigure[TransR]{\includegraphics[width=0.7\textwidth]{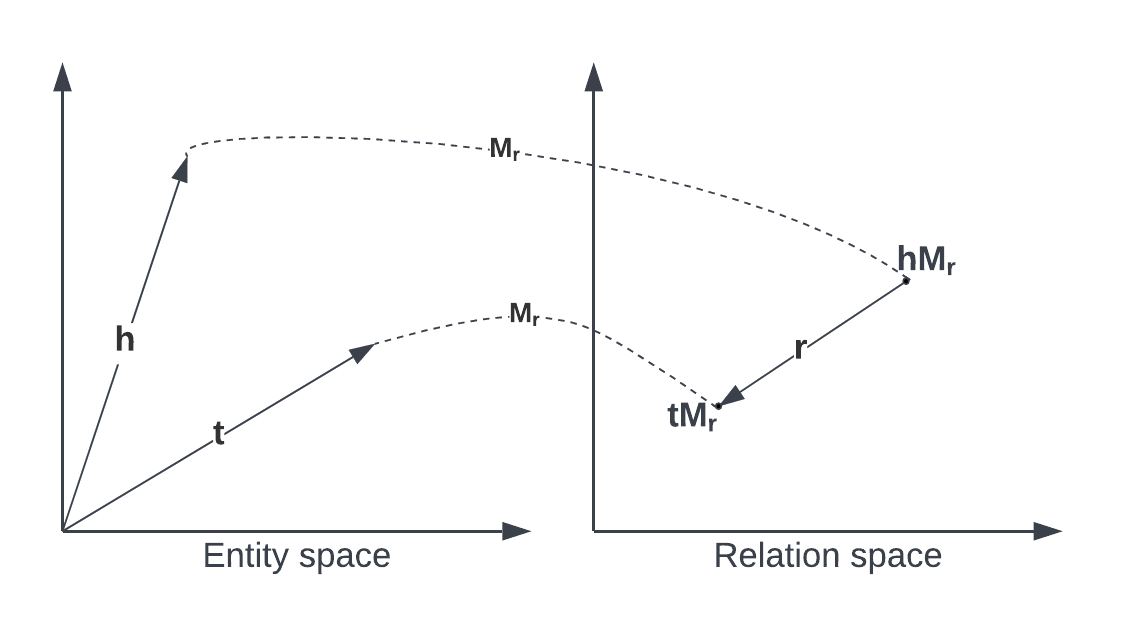}\label{fig:transR}}
    \label{fig:translation-based models}
    \caption{Simplified illustrations of entities and relations in TransE, TransH, and TransR.}
\end{figure*}

\begin{itemize}
    \item \textbf{TransE} \cite{bordes2013translating} is a translation-based model that learns low-dimensional embeddings of entities and represents relationships as translations in the embedding space. The objective of TransE is to minimise the distance between vectors $\mathbf{h+l}$ and $\mathbf{t}$ if the triple $(h,r,t)$ holds, or to maximise the distance conversely, as described in Figure \ref{fig:transE}. Accordingly, the scoring function of TransE can be represented using Equation \ref{eq:transE}. Although TransE can effectively handle 1-to-1 relations in the knowledge graph, it still remains flaws for 1-to-N, N-to-1, and N-to-N relations.
    \begin{equation}\label{eq:transE}
         f_r(\mathbf{h},\mathbf{t})=-\parallel\mathbf{h}+\mathbf{r}-\mathbf{t}\parallel_2^2
    \end{equation}

    \item \textbf{TransH} \cite{wang2014knowledge} overcomes the problems of TransE in modelling 1-to-N, N-to-1, and N-to-N relations by enabling entities to have distributed representations in different relations. Specifically, as described in Figure \ref{fig:transH}, TransH first introduces a hyperplane $\mathbf{w}_r$ to represent a specific relation $r$ and positions the translation vector $\mathbf{d}_r$ in the hyperplane. Then it projects $\mathbf{h}$ and $\mathbf{t}$ to the hyperplane $\mathbf{w}_r$, denoted by $\mathbf{h_\perp}$ and $\mathbf{t_\perp}$, respectively. The expectation is that $\mathbf{h_\perp} + \mathbf{d}_r$ approaches $\mathbf{t_\perp}$ if $(h,r,t)$ holds. Equation \ref{eq:transH} formulates the scoring function of TransH.
    \begin{equation}\label{eq:transH}
        f_r(\mathbf{h},\mathbf{t})=-\parallel \mathbf{h_\perp}+\mathbf{d_r}-\mathbf{t}_\perp\parallel_2^2
    \end{equation}
    where $\mathbf{w}_r$ is the normal vector, $\mathbf{h_\perp} = \mathbf{h}-\mathbf{w}_r^{\top}\mathbf{h}\mathbf{w}_r$, and $\mathbf{t}_\perp = \mathbf{t}-\mathbf{w}_r^{\top}\mathbf{t}\mathbf{w}_r$. Through this mechanism, TransH enables entities to have diverse roles in different relations.

    \item \textbf{TransR} \cite{Lin2015Learning} further improves the embedding performance by modelling entities and relations into distinct embedding spaces because entities and relations are completely different objects. To perform the translation in such settings, TransR sets entities embeddings as $\mathbf{h}, \mathbf{t}\in \mathbb{R}^k$ and the relation embedding as $\mathbf{r} \in \mathbb{R}^d$ for any triple $(h,r,t)$, where $k\neq d$. Meanwhile, a projection matrix $\mathbf{M}_r \in \mathbb{R}^{k\times d}$ is used to project entities from the entity embedding space into the relation embedding space, as shown in Figure \ref{fig:transR}. The scoring function of TransR is described in Equation \ref{eq:transR}.
    \begin{equation}\label{eq:transR}
        f_r(\mathbf{h},\mathbf{t})=-\parallel \mathbf{h}\mathbf{M}_r+\mathbf{r}-\mathbf{t}\mathbf{M}_r\parallel_2^2
    \end{equation}
\end{itemize}

\subsection{Dual Observations}
In the current setting, the dual observation aims to access and refine the user's reading preference by considering both the focus of the news article and the user's belief network. It consists of internal observation and external observation. 

The internal mechanism combines low-order relation representations with high-order relation expressions, which helps to avoid data leakage and the cold-start problem and allows for the exploration of more meaningful information for the user. The external observation mechanism focuses on the continual influence of each piece of news on the user. Every time a user reads a piece of news, their belief network is connected to the primary semantic information and historical knowledge of the news due to the mutual influence between each word in the news and the influence between the user's belief network and the news. Different users may show different levels of attention to the same article. Hence, the observation of internal news features and the interaction between the user's belief network and the news is essential for extracting the user's current knowledge system.

Dual observation is different from dual attention, which generally refers to the use of two separate attention mechanisms in a single model, where the two attention mechanisms can be used independently or in combination to attend to different aspects of the input data. For example, a model with dual attention might use one attention mechanism to focus on user information and another attention mechanism to focus on item information when making recommendations \cite{jiang2022san}. By contrast, dual observation emphasises the vital information of the article and the user's belief network.

\begin{figure}
		\centering
  \caption{A typical process of how readers perceive and understand information.} 
		\includegraphics[width=1.0\textwidth]{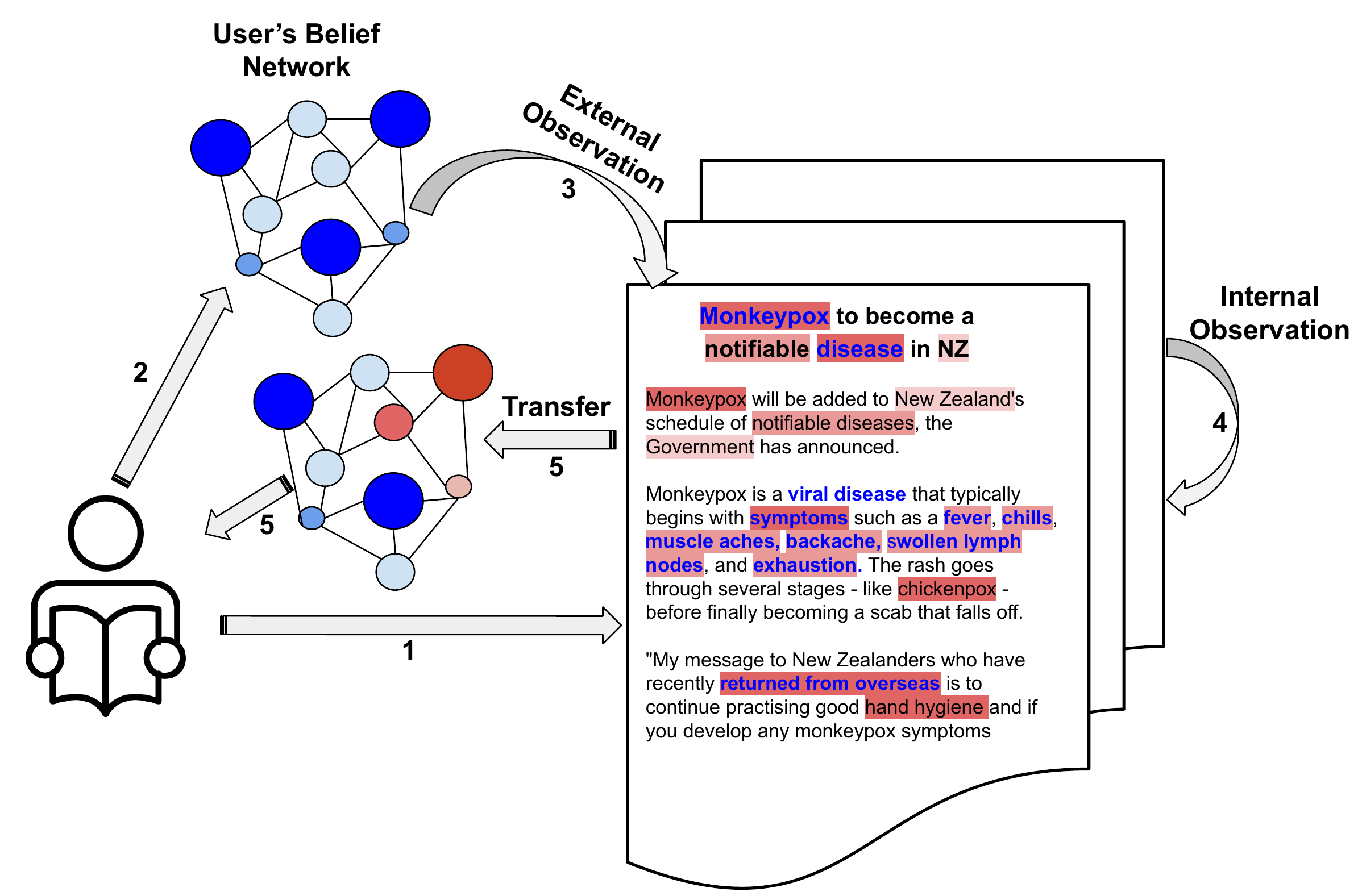}
		\label{fig:combination}
	\end{figure}

Figure \ref{fig:combination} demonstrates how readers perceive and understand information. Users form their knowledge system mainly through two observation mechanisms. The first is the observation of the internal focus of the news, which entirely extracts the news feature (the word with red colour, with darker red representing more importance). When a user read this article, they will analyse and perceive the information based on the existing belief network, describing their prior knowledge and experiences. The pure blue user belief network (BUBN) represents the user's prior knowledge. The second observation mechanism is the external observation, which observes the mutual influence between the BUBN and the read article. The final affected result (mixed colour user belief network) is transferred back to the user. It can be seen that when a person reads a text, they not only receive the information of the article but also incorporate it into their existing belief networks. It is vital to adopt dual observation mechanisms in order to retrieve users' preferences accurately.

\subsection{Problem Definition} 
In this research work, we focus on investigating a novel personalised news recommendation based on users' beliefs and news features. Formally, we define a reader $r$'s reading behaviour as $b_{r,t} \in B_r$, where $B_r$ is a list of reader $r$'s historical reading behaviours. $b_{r,t}$ is a four-tuple, i.e., $b_{r,t}=(reader_{id}, news_{id}, t, l)$, where $reader_{id}$ represents the reader id, $news_{id}$ represents the news id linking to a piece of specific news, $t$ implies to the time stamp, and $l$ is the label indicating the reading status, either clicked news or non-clicked news. Each news consists of a title and an abstract, which can be represented as a sequence of words $[w_1, w_2, \dots, w_m]$. The final purpose of our research is to distinguish whether the reader r will select a piece of candidate news and calculate the click probability.

\section{Dual-Observation based Recommendation}
	
In this section, we will thoroughly explain the proposed Dual-Observation based Recommendation (DOR) model by first outlining its architecture. We will then delve into the intricacies of the dual-observation modules, explicitly focusing on the internal and external observation mechanisms.
	
\subsection{DOR Architecture}
	
\begin{figure}
		\centering
  
		\includegraphics[width=1.0\textwidth]{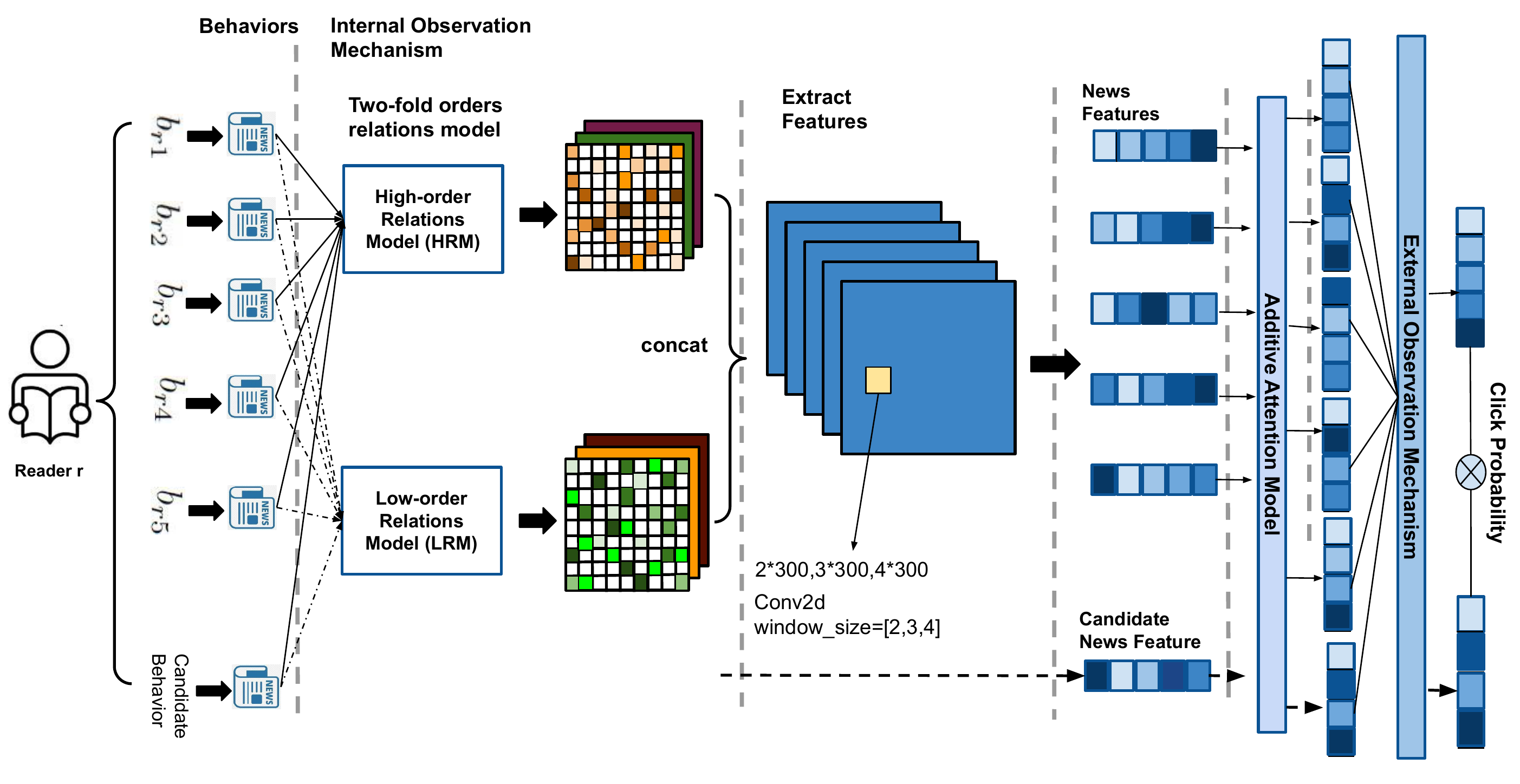}
  \caption{The overall architecture of DOR.} 
		\label{fig:DOR}
	\end{figure}

	
The overall architecture of the proposed DOR model is demonstrated in Figure \ref{fig:DOR}. The model primarily comprises two main modules: the internal and external observation mechanisms. 
	
	
As illustrated in Figure \ref{fig:DOR}, a reader $r$ reading histories $B_r$ = $\{b_{r1}, b_{r2}, b_{r3}, b_{r4}, b_{r5}\}$ are fed into the DOR system. The internal observation mechanism simultaneously processes these inputs by utilising the high-order relation model (HRM) and the low-order relation model (LRM). These two relations models learn news features from contextual and content perspectives, respectively. The HRM module incorporates external knowledge from knowledge graphs to overcome the limitations of information sparsity and learns contextual news representations at a high-order level. Meanwhile, to extract rich news features, the LRM module is used within the internal observation mechanism to obtain lexical-level representations. The high-low bipartite representations of news are concatenated and then trained by the feature extractor. The internal observation mechanism then observes the internal weights, using an additive attention model, to differentiate between different attentions among entities and words for the user. On the other hand, the external observation mechanism delves into the deep relationships between the user's evolving belief and the input news. As the user receives new information, the corresponding belief is continuously changing. The external observation mechanism examines the significant mutual influence between the user's belief and the input news. Lastly, similarity calculation is used to calculate the probability of the user clicking on the candidate news.

The details of the internal and external observation mechanisms are given in the following subsections.

\subsection{High-Order and Low-Order Relations}

The internal observation module concentrates on both the content and contextual information of each news item. It aims to identify the inherent characteristics of news articles by utilising both high and low-level representations. The readers' perception of the information is partially influenced by the main idea of the reading materials, where different words are assigned varying levels of importance for the news.

As depicted in Figure \ref{fig:DOR}, the internal observation mechanism includes two essential models, i.e., the low-order relations model (LRM) and the high-order relations model (HRM). The LRM captures the lexical-level representations of news articles, and the HRM incorporates external knowledge from knowledge graphs to extract contextual news representations at a high-order level. These two models work in tandem to provide a comprehensive understanding of the characteristics of news articles. Next, the LRM and HRM are introduced with examples. 

\begin{figure}
		\centering
		\includegraphics[width=1.0\textwidth]{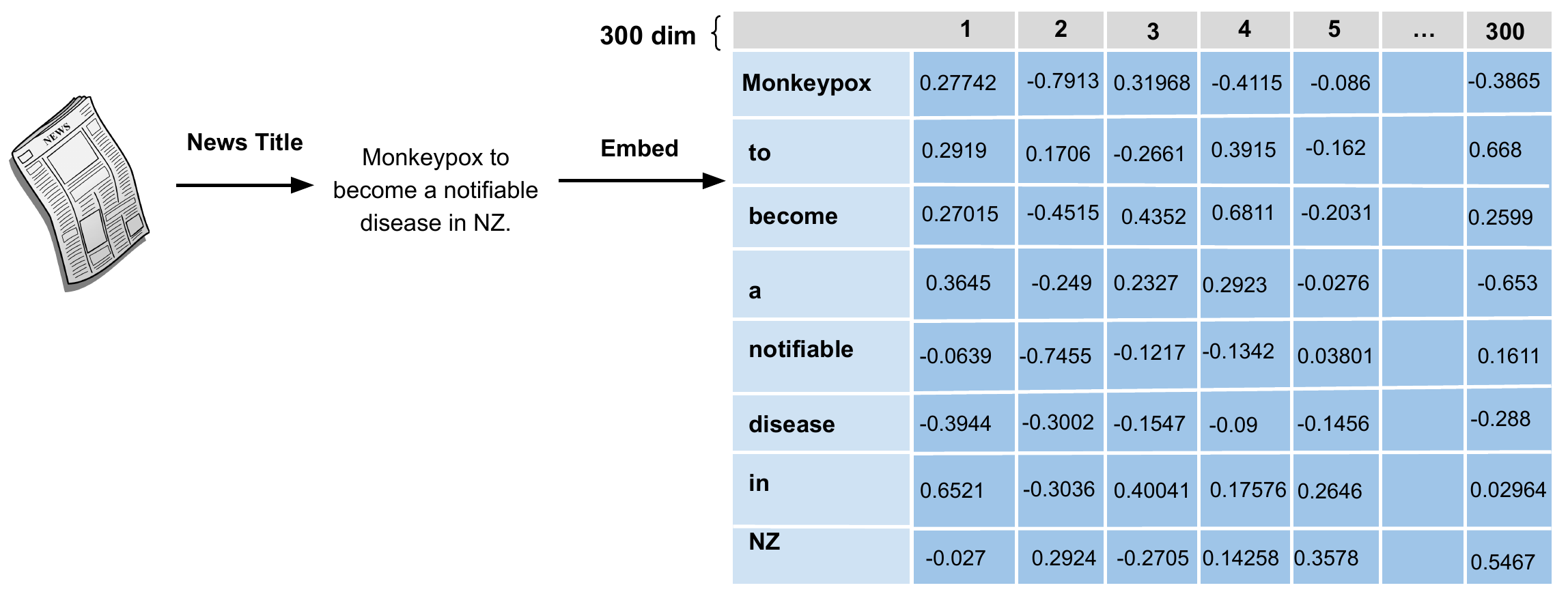}
    \caption{An Example of Low-order Relations Model (LRM).}
	 \label{fig:LRM}
   \end{figure}

    In the current setting, the low-order relation captures the relationships between the words within a news article. Each word in the article is represented as a vector, and these vectors are used to depict the connections between the words. The LRM module aims to obtain the lexical-level representations of the news articles by representing the relationships between the words, and such representations are used for further analysis. With this mechanism, the DOR system can acquire the internal information of the news articles and use them for recommendations. An example of the LRM module is illustrated in Figure \ref{fig:LRM}. Taking the news title "Monkeypox to become a notifiable disease in NZ" as an input, the textual title is passed through a pre-trained word embedding model to acquire the title's representations.

\begin{figure*}
		\centering
		\includegraphics[width=1.0\textwidth]{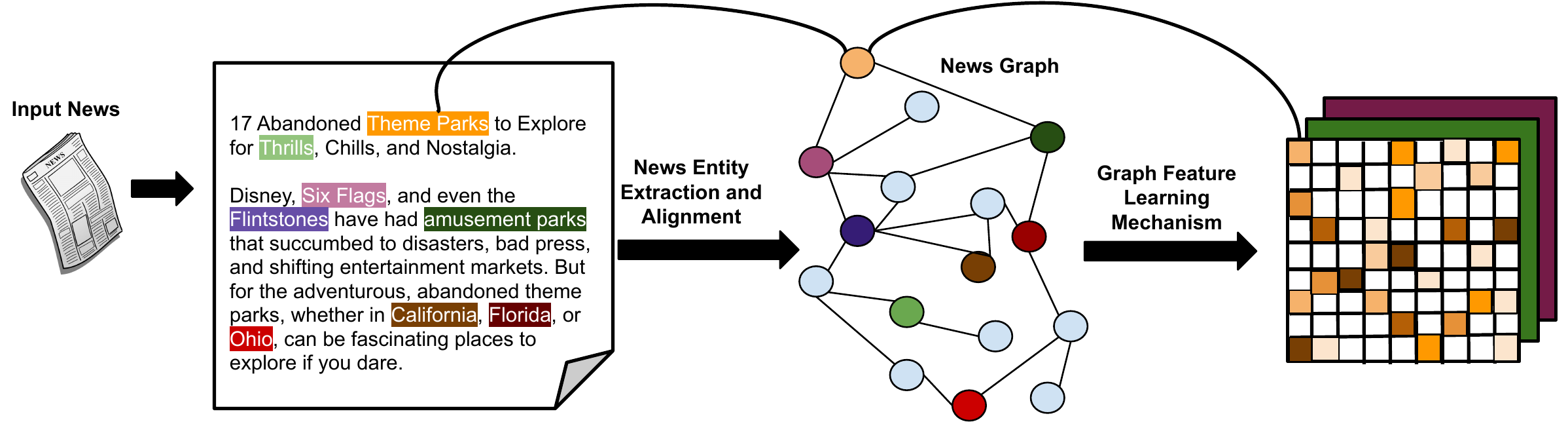}
         \caption{An Example of High-order Relations Model (HRM).}
	 \label{fig:HEM}
	\end{figure*}

    On the other hand, high-order relations describe the connections between entities mentioned in the news article that are not explicitly mentioned in the text but can be inferred through additional context or external knowledge. To identify these connections, it uses techniques such as knowledge distillation and entity linking \cite{milne2008learning}, which align entities present in the news content with pre-defined entities in an external knowledge graph. In this research, we use Wikidata \footnote{https://www.wikidata.org/} as the external knowledge graph, which is a vast repository of structured data from the real world. To further illustrate HRM, an example is presented in Figure \ref{fig:HEM}. The entities ``Theme Parks'' and ``Thrills'' are distilled and aligned with the corresponding entities (in yellow and light green, respectively) in the knowledge graph. With the assistance of external knowledge, HRM can provide a contextual understanding of the news article, which will later be used for recommendations.

\subsection{Graph Feature Learning}

    Graph feature learning is utilised to extract the neighbours and relations of entities in a news-domain graph, which is then transferred into a graph-based feature learning mechanism (GFLM). The GFLM discovers the node's embedding by enriching contextual information, including multiple-head neighbour expressions and their corresponding relations' representations. The GFLM further enhances the representation of news by incorporating an external knowledge graph and considering the contextual information within the news-domain graph. Additionally, the HRM model accentuates the diverse importance of each node by the GFLM, which allows an understanding of the significance of each entity better.

    Inspired by previous works in this area (such as \cite{nathani2019learning}), the GFLM is employed to learn nodes' representations from a high-order level. The GFLM is a novel neural network architecture that deals with graph-structured data. It utilises masked self-attention layers to tackle the limitations of earlier graph convolution-based methods. In the standard signal layer of GFLM, the inputs include a set of entities' features, $E =\{\mathbf{e}_1,\dots,\mathbf{e}_i\}$, where $\forall\mathbf{e}_i\in \mathbb{R}^d$. The outputs of GFLM indicates as $E' = \{\mathbf{e}'_1,\dots,\mathbf{e}'_i\}$, where $\forall\mathbf{e}'_i\in \mathbb{R}^{d'}$. In the current setting, the head $h$ and tail $t$ of a fact are seen as the entity $e$ in the GFLM model. We convert $h$ and $t$ to $\mathbf{h}$ and $\mathbf{t}$, which are the feature of $h$ and $t$. The resultant of $\mathbf{h}$ and $\mathbf{t}$ is supposed to be $\mathbf{e}$, which are learned using translation-based representation learning models, such as TransE, TransH, and TransR models. Furthermore, each target entity embedding $\mathbf{e}$ is represented by 2-hop neighbour entities' and relations' embeddings with attention. The same approach is applied to $\mathbf{h}$ and $\mathbf{t}$. As an original step, a shared linear transformation is provided to each node and parametrised by a weight matrix $\mathbf{W}$ $\in$ $\mathbb{R}^{{d'} \times d} $. An attention mechanism $a$ : $\mathbb{R}^{d'} \times \mathbb{R}^{d'}\xrightarrow{1}\mathbb{R}$ is applied to compute attention coefficients. For example, given the importance of $\mathbf{e}_i$ to $\mathbf{e}_j$ as $\mathbf{e}_{ij}$, the coefficient of attention between $\mathbf{e}_i$ and $\mathbf{e}_j$ can be calculated by using the following equation:

	\begin{equation}
		\mathbf{e}_{ij} = a(\mathbf{W}\mathbf{e}_i, \mathbf{W}\mathbf{e}_j).
		\label{eq:coefficient_attention}
	\end{equation}

	The attention mechanism used in the experiments is a type of neural network architecture known as a single-layer feed-forward neural network. The attention mechanism is parameterised, meaning it has a set of adjustable parameters that can be learned during training, represented by a weight vector $\bm{a}\in \mathbb{R}^{2d'}$. A $softmax$ function is then applied to the attention coefficients to enable easy comparison among different entities. The non-linear activation function ${LeakyReLU}$ with a 0.2 negative input slope $\alpha$ value is adopted to the algorithm. Finally, the attention coefficient $\alpha_{ij}$ can be computed using the following equation:
		\begin{equation}
		\alpha_{ij} = \frac{\exp(\sigma(\mathbf{e}_{ij}))}{\Sigma_{k\in{N_i}}\exp(\sigma(\mathbf{e}_{ik}))}
		= \frac{\exp(LeakyReLU(\bm{a}^\top[\mathbf{W}\mathbf{e}_i\Vert\mathbf{W}\mathbf{e}_j]))}{\Sigma_{k\in{N_i}}\exp({LeakyReLU}(\bm{a}^\top[\mathbf{W}\mathbf{e}_i\Vert\mathbf{W}\mathbf{e}_k]))}, 
	\end{equation}
 \noindent

	In order to improve the stability of GFLM, a multi-head GFLM method is utilised. This method is formulated using the following equations:
	
	\begin{equation}
		\mathbf{e}'_i = \sigma (\sum_{j\in{N_i}}\alpha_{ij}\mathbf{W}\mathbf{e}_j),
	\end{equation}
	
	\begin{equation}
		\mathbf{e}'_i = {\mathbin\Vert_{k=1}^{K}} \sigma (\sum_{j\in{N_i}}\alpha^k_{ij}\mathbf{W}\mathbf{e}_j),
	\end{equation}
	
	\begin{equation}
		\mathbf{e}'_i = \sigma (\frac{1}{K}\sum_{k=1}^{K}\sum_{j\in{N_i}}\alpha^k_{ij}\mathbf{W}\mathbf{e}_j),
	\end{equation}
where $N_i$ indicates the neighbourhoods of node $e_i$ in the graph. $\parallel$ means the concatenation operation. The relations significantly contribute to understanding the graph and context \cite{nathani2019learning}. The DOR incorporates relation features into the representations of contextually related entities. The edge between node $e_i$ and node $e_j$ is denoted as $\hat{e}_{ij}$. As previously mentioned that the entity representation is represented by the $E\in\mathbb{R}^{K}$ matrix, and then we define that the relation representation is represented by a matrix $R\in\mathbb{R}^{T}$. $T$ is similar to the $K$, which means the feature dimensions of each entity representation. Thus, the $\mathbf{e}_{ij}^{rel}$ can be represented by the matrices of $E'\in\mathbb{R}^{K'}$ and $R'\in\mathbb{R}^{T'}$as:

	\begin{equation}
		\mathbf{e}_{ij}^{rel} = a(\mathbf{W}\mathbf{e}_i, \mathbf{W}\mathbf{e}_j, \mathbf{W}\mathbf{\hat{e}}_{ij}),
	\end{equation}
	
	The targeted node representation can be described via the neighbours' embedding with corresponding relations' embedding. The embedding of entity $e'_i$ is as below:
	
	\begin{equation}
		\mathbf{e}'_i = \sigma (\frac{1}{K}\sum_{k=1}^{K}\sum_{j\in{N_i}}\sum_{k\in{R_{ij}}}\alpha^k_{ij}\mathbf{W}\mathbf{e}_j),
	\end{equation}
	where $K$ represents the number of hops. In the current settings, we incorporate 2-hop neighbour entities inside the target entity embedding, i.e., $K=2$. $\alpha$ denotes the attention coefficient and $\sigma$ refers to the activation function, i.e., LeakyReLU. 
	
 
	The embedding fusion strategy has been applied to fuse both embeddings, shown in Figure \ref{fig:DOR}. A news article is composed of multiple entities represented as $\mathbf{N}$. Therefore, the news can be represented as $\mathbf{N}_1, \mathbf{N}_2,...,\mathbf{N}_m$, where $m$ represents the number of entities in the news. A CNN model is adopted to further learn the representation of the news as a whole
 .
	
	\begin{equation}
		\mathbf{N}_j = \mathbf{W_{N}}_j \cdot output(GFLM(TransR({N_{j}}))),
	\end{equation}

    In the DOR method, the Input News Section processes each piece of news and extracts its features using a CNN model. To further enhance these features, an Additive Attention Mechanism (AAM), which is an attention-based deep neural network, is used in the internal observation mechanism. As a result, the final representation of each piece of news, denoted as $\bm{N}_{final}$, can be obtained by the following equation:

	\begin{equation}
		\mathbf{N}_{final} = AAM(\mathbf{N}_j),
	\end{equation}

	\subsection{Knowledge Graph Distillation and Construction}

    In the proposed DOR, three approaches are applied to address the issue of inadequate entities and relations distillation. Firstly, it utilises both the news title and the news abstract as inputs to the DOR, thereby revealing the accurate meaning of the news. Secondly, it employs GloVe2vec to embed all words of each news from the perspective of low-order relations. Thirdly, it incorporates an external knowledge graph to enhance the representation of entities in the news.

\begin{figure}
		\centering
  \caption{An Example of News Knowledge Graph Construction and Representation.}
		\includegraphics[width=1.0\textwidth]{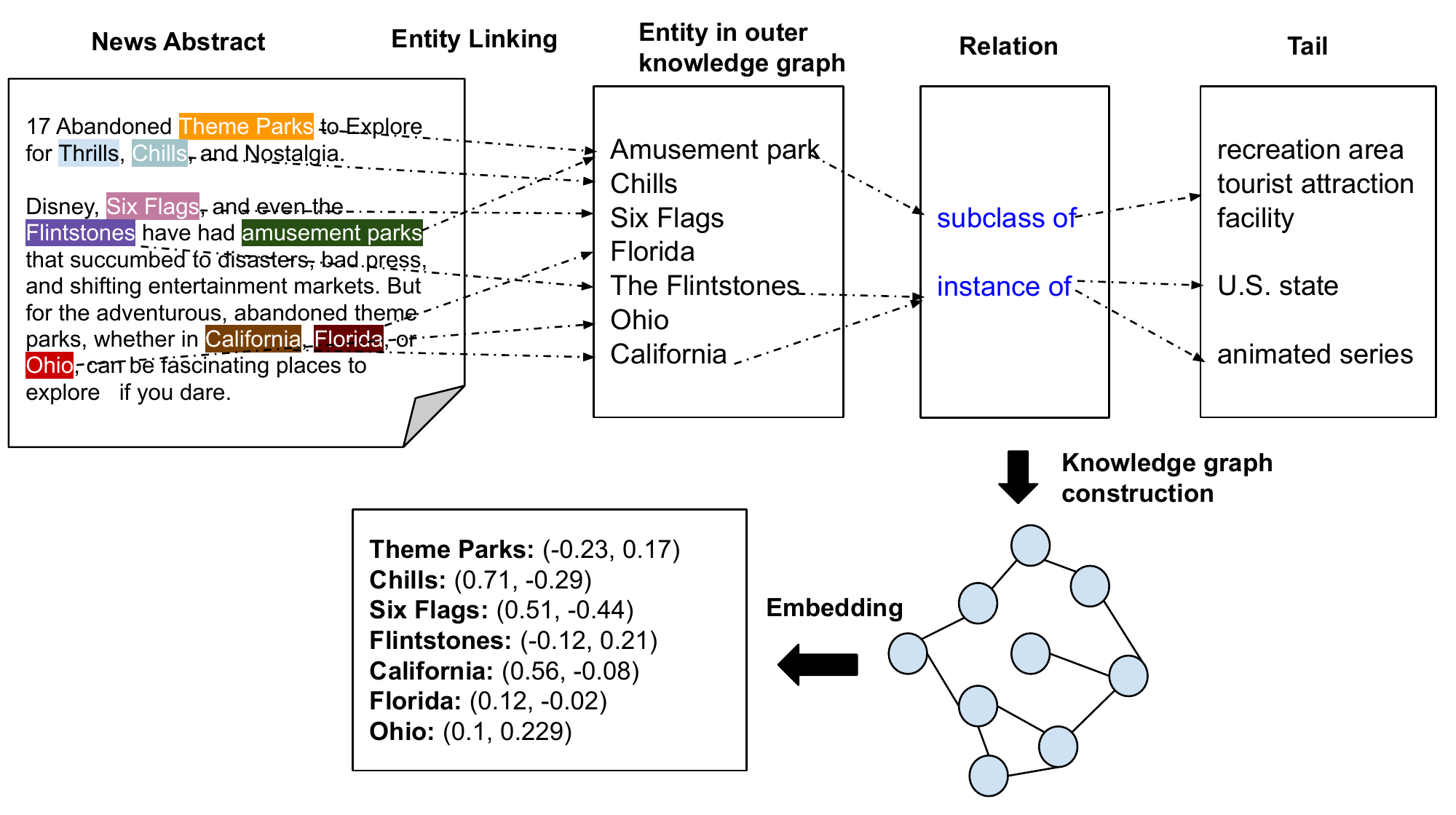}
	 \label{fig:KGConstruction}
	\end{figure}

    An example of the news knowledge graph extraction and construction process is illustrated in Figure~\ref{fig:KGConstruction}. The input is a text composed of a news title and abstract. Entities are extracted from the text and linked with corresponding entities in the Wikipedia knowledge graph. Based on these entities, their neighbours and relations are also explored. All the extracted information is then used to construct a knowledge graph of news.

	
	\subsection{External-Observation Mechanism}

    The External Observation Mechanism (EOM) in the DOR system receives the content and contextual features of the news from the internal observation mechanism. The focus of this module is to study the different influences on the user's behaviour from each piece of news. Therefore, the EOM is a deep learning model with a self-attention mechanism that analyses the mutual influence between users' beliefs and each piece of news. This module allows investigation of how users' perceptions and behaviours are affected by the news.
 
	Specifically, the EOM takes into account the dynamic nature of users' beliefs. It incorporates a sequential characteristic that considers each historical reading record as a snapshot of the user's evolving belief network. The EOM receives the representations of the current news from the internal observation mechanism and uses the user's current belief network to output the user's preferences. Thus, it models how the user's perception of current news is influenced by their previous reading history and belief network, which makes our model more adaptive to the user's dynamic interests.
	
    Mathematically, given user $i$ embedding $U_{i_e}$ and the candidate news embedding $N_{i_e}$, the probability of user $i$ clicking news $N_i$, i.e., $p_{i,N_i}$, is estimated through a general deep neural network $\mathbf{D}$:
	
	\begin{equation}
		p_{i,N_i} = \mathbf{D}(U_{i_e}, N_{i_e})
	\end{equation}

\section{Experiment and Analysis}

    In this section, we first evaluate the performance of the proposed DOR model using two real-world datasets by comparing it against a few popular baselines. Subsequently, an ablation study is conducted to verify the contribution of the dual observation mechanisms. Finally, we discuss the impact of the chosen parameters on our model's performance.

\subsection{Dataset Description}
    In all of our experiments, we utilise the Microsoft News datasets (MIND)\footnote{https://msnews.github.io/}, publicly available and large-scale news recommendation datasets. We adopt two versions of the datasets, namely the MIND-small and MIND-large versions, which include user reading behaviours of 50,000 and 711,222, respectively. The statistics of both datasets are listed in Table \ref{table:datasets}. 
    
    As can be observed from both tables, the user reading behaviours in the MIND-large datasets are significantly higher than the MIND-small datasets. Consequently, the total number of words and entities is vastly different in these two datasets. The MIND-small datasets comprise approximately 70,000 words and 25,000 entities, while the MIND-large datasets have approximately 100,000 words and 21,000 unique entities.

\begin{table}[t]
\centering
\caption{Statistics of datasets}
\label{table:datasets}
\begin{tabular}{l|ll}
\hline
\textbf{Dataset}                        & \textbf{MIND-small} & \textbf{MIND-large} \\ \hline
\textbf{No. users}                      & 50,000              & 711,222             \\
\textbf{No. behaviours}                 & 230,117             & 2,232,748           \\
\textbf{No. words}                      & 70,975              & 101,232             \\
\textbf{No. entities}                   & 25,916              & 21,843              \\
\textbf{Maximum No. words per title}    & 20                  & 20                  \\
\textbf{Maximum No. words per abstract} & 50                  & 50                  \\ \hline
\end{tabular}%
\end{table}

    The datasets are pre-processed by removing news items with fewer than 10 readers and filtering out readers with fewer than 50 reading records. Furthermore, the datasets are integrated with users' reading information, including user ID, news title, title entities, abstract, abstract entities, impressions, and labels (0 for non-click and 1 for click). Both datasets are partitioned into training and testing sets in a ratio of 80:20, respectively. Each news item in the dataset is limited to 20 words for the title and 50 words for the abstract.

\subsection{Experimental Set-up and Implementations}

    We adopt PyTorch to implement all models with the following settings. The epoch size is set to different values for both datasets, which are 5, 8, 10, 12, and 15. The batch size is set as 128, and the learning rate is defined as 0.0001. The word dimension in our experiment is set to 300, and the context embedding of each entity is also set to 300. The pre-trained GloVe2Vec \cite{pennington2014GloVe} method is used as the word embedding in our work. To evaluate its performance, well-known embedding mechanisms, such as Word2Vec \cite{church2017Word2Vec} and BERT \cite{reimers2019sentence}, are also employed in our experiments. 
    
    On top of that, to enhance the graph extracted from news, we utilise Wikidata \footnote{https://www.wikidata.org/} to augment the graph representation with different entity neighbour sizes, i.e., 5, 10, 15, and 20. For example, for a given neighbour size of 5, each news entity forms an ego-network through Wikidata with a node degree of 5. As the neighbour size increases, the resulting augmented graph also expands in size. The integration of these ego-networks forms the final augmented graph that is used in our model.

    The DO mechanism of our model extracts the context embedding and is trained with a batch size of 110. The dimension of context embedding for each entity is set to 300. Furthermore, the input for the DO module is pre-processed using the TransR model, which has 300 dimensions and is trained using 10 batches. The DOR model is trained using the Adam optimization method \cite{kingma2014adam} through log loss optimization.

\subsection{Evaluation Metrics}

In evaluating the performance of our model, we use several commonly used metrics in the field of recommendation systems \cite{krichene2022sampled}.

\begin{itemize}
    \item AUC (Area under the ROC Curve): AUC is a measure of the probability that randomly chosen related items will rank higher than randomly chosen unrelated items. A higher AUC value indicates that the model is better at distinguishing between related and unrelated items.
    
    \item MRR (Mean Reciprocal Rank): MRR indicates the mean value of the reciprocal rankings of multiple query statements. It is a measure of the effectiveness of a ranking system, with a higher MRR indicating a higher level of effectiveness.
    
    \item NDCG (Normalised Discounted Cumulative Gain): NDCG is a measure of the ranking quality of a recommendation system. The principle of NDCG is that highly correlated products should rank higher than unrelated products. A higher NDCG value indicates a better ranking of related products.
    \subitem NDCG@5: The NDCG@5 metric calculates the DCG of the first five recommendations.
    \subitem NDCG@10: The NDCG@10 metric calculates the DCG of the first ten recommendations.
\end{itemize}

These metrics allow us to evaluate the effectiveness of our model in different aspects, such as its ability to rank relevant items higher than irrelevant items, and its ability to rank the most relevant items higher than the less relevant items. A high score on these metrics indicates the model is able to generate relevant and accurate recommendations.
	
\subsection{Baseline Methods}

In order to evaluate the performance of our model, we compare it against several popular baselines in the field of news recommendation. These baselines are:

\begin{itemize}
    \item \textbf{DFM} \cite{lian2018towards} is a deep combination architecture that combines several layers with different depths.
    \item \textbf{DKN} \cite{wang2018dkn} is a news recommendation system that adopts attention networks to learn the representation of the entity-word level. The model proposed in DKN uses a dual-attention mechanism.
    
    \item \textbf{Hi-Fi Ark} \cite{liu2019hi} summarises reader histories into vectors as documents and studies candidate-dependent reader expressions through the careful aggregation of such documents.
    
    \item \textbf{TANR} \cite{wu2019bneural} captures topic-aware representation as the news encoder and user-aware representation as the user encoder. The model proposed in TANR uses attention-based news expression as the news encoder and user-aware expression as the user encoder.
    
    \item \textbf{FIM} \cite{wang2020fine} is a fine-grained interest-matching model adapted for neural news recommendation.
    
    \item \textbf{NRMS} \cite{wu2019neural} proposes an approach for insufficient user interest extraction based on the news title. The model proposed in NRMS considers both the news title and news abstract for building reading interest.
\end{itemize}

\subsection{Performance Evaluation}

\begin{table}[]
\begin{center}
\caption{Performance Evaluation using MIND-Small datasets}\label{tab:mind_small}%

\begin{tabular}{lllll}
\hline
\textbf{Model} & \textbf{AUC}   & \textbf{MRR}   &  \textbf{NDCG@5} & \textbf{NDCG@10} \\ \hline
DKN            & 58.26          & 25.76          & 27.18                                        & 34.19                                         \\
DFM            & 59.96          & 31.13          & \textbf{32.86}                               & 41.65                                         \\
Hi-Fi Ark      & 60.16          & 30.62          & 32.25                                        & 41.14                                         \\
TANR           & 61.56          & 27.46          & 29.81                                        & 36.49                                         \\
NRMS           & 62.78          & 29.4           & 31.61                                        & 38.78                                         \\
FIM            & 63.29          & 30.65          & 33.71                                        & 40.16                                         \\
DOR            & \textbf{64.35} & \textbf{33.39} & 32.68                                        & \textbf{41.78}                                \\ \hline
\end{tabular}
\end{center}
\end{table}

\begin{table}[]
\begin{center}
\caption{Performance Evaluation using MIND-Large datasets}\label{tab:mind_large}%
\begin{tabular}{lllll}
\hline
\textbf{Model} & \textbf{AUC}   & \textbf{MRR}   & \textbf{NDCG@5} & \textbf{NDCG@10} \\ \hline
DKN            & 62.42          & 30.22          & 32.93           & 38.46            \\
DFM            & 60.29          & 32.22          & 32.98           & \textbf{42.86}   \\
Hi-Fi Ark      & 62.23          & 31.16          & 32.34           & 42.65            \\
TANR           & 64.62          & 30.89          & 33.84           & 40.35            \\
NRMS           & 64.98          & 32.37          & 35.82           & 42.34            \\
FIM            & 65.36          & 33.66          & 35.35           & 42.02            \\
DOR            & \textbf{65.72} & \textbf{33.89} & \textbf{35.84}  & 42.35            \\ \hline
\end{tabular}
\end{center}
\end{table}

In order to evaluate the performance of our model, we compare it against several popular baselines in the field of news recommendation, as previously mentioned. The experimental results of these models are depicted in Tables \ref{tab:mind_small} and \ref{tab:mind_large}. All models shown in these tables are state-of-the-art news recommendation architectures and they all consider the importance of attention for constructing more satisfactory user representations. However, the same data can have different expressiveness in different news recommendation neural networks. Unlike other baselines, our proposed DOR model presents a Binary Observations (BO) mechanism-based recommendation system, which aims at constructing user preference by considering more detailed contextual expressions. As can be seen from Table \ref{tab:mind_large}, all models perform better when trained using the MIND-large datasets. The performance of these models is described through four evaluation matrices, and our DOR model outperforms the other six baselines in all of them. This proves the superior stability and precision of our DOR model.

In contrast, the DKN model has unremarkable results on all evaluation matrices and gets the lowest score of 25.76 on the MRR normalisation. In addition, for the NDCG scores, the DFM, Hi-Fi Ark, and FIM approaches outperform other baselines, obtaining scores over 40, indicating the relatively high accuracy of these models. For example, the DFM model achieves 42.86 points on the NDCG@10 standard when it employs larger datasets. However, our model only gets 42.35 NDCG@10 scores with the larger datasets but outperforms when using the smaller datasets. Overall, our proposed model has better performance in terms of both stability and accuracy.

 \subsection{Ablation Study}	

\begin{table}[t]
\begin{center}
\caption{Ablation Study on Dual Observation.}\label{tab:DO}%
\begin{tabular}{lllll}
\hline
\textbf{Model} & \textbf{AUC}   & \textbf{MRR}   & \textbf{NDCG@5} & \textbf{NDCG@10} \\ \hline
With DO           & 64.35          & 33.39          & 32.68           & 41.78           \\
Non-DO            & 58.36          & 25.88          & 28.16           & 35.17   \\
DO with L     & \textbf{65.72}       & \textbf{33.89}          & \textbf{35.84}           & \textbf{42.35}            \\
Non-DO with L           & 62.67          & 27.95          & 31.37           & 38.16            \\
\hline
\end{tabular}
\end{center}
\end{table}

\begin{table}[t]
\begin{center}
\caption {Study on Low-order Relation Representations.}\label{tab:LORR}
\begin{tabular}{lllll}
\hline
\textbf{Model} & \textbf{AUC}   & \textbf{MRR}   & \textbf{NDCG@5} & \textbf{NDCG@10} \\ \hline
Word2Vec          & 60.25          & 26.57          & 28.66           & 35.26           \\
GloVe            & 64.35          & 33.39          & 32.68           & 41.78   \\
BERT     & 59.47       & 25.71         & 27.78           &34.34            \\
Word2Vec\textit{w}L          & 62.28          & 31.89          & 29.84           & 35.25            \\
GloVe\textit{w}L          & \textbf{65.72}          & \textbf{33.89}         & \textbf{35.84}           & \textbf{42.35}            \\
BERT\textit{w}L          & 61.01         & 27.29          & 29.2           & 35.22            \\
\hline
\end{tabular}
\end{center}
\end{table}
To evaluate the contributions of modules in the DOR model, we conduct three ablation experiments using two datasets. The first ablation experiment examines the impact of our proposed Dual Observations (DO) mechanism. To evaluate the contribution of DO, we compare the performance of DOR with and without DO. The second ablation experiment examines the effectiveness of Low-order Relation Representations. We use pre-trained GloVe word representations in our model to meet our requirements for low-order relation representations. In this ablation experiment, we replace the GloVe model with two other pre-trained models, Word2Vec and BERT, to evaluate their performance on our model. The third ablation study examines the impact of Diverse Translational Distance Models (DTDMs), such as TransE, TransH, and TransR on our model.

According to Table \ref{tab:DO}, it is clear that the DO mechanism plays a crucial role in our approach. The performance of our model with the DO mechanism is significantly better than that of the model without it. As can be seen from Table \ref{tab:DO}, whether it be model stability matrices or model accuracy matrices, our proposed model improves the performance of the system. L in Table \ref{tab:DO} indicates the larger datasets (MIND-large datasets) in our experiments. All values with the larger datasets in Table \ref{tab:DO} are more significant than the values with the smaller datasets. The model incorporating the DO mechanism and utilising a larger database delivers exceptional performance according to evaluation criteria.

Table \ref{tab:LORR} illustrates the different performances among three ablation models using two distinct datasets. Similarly, Table \ref{tab:LORR} also displays that larger datasets have more expressive than light datasets on our model. It is evident that the GloVe model consistently outperforms the other models in all evaluation matrices, and our model achieves about 65 AUC degrees with the GloVe model based on the more extensive datasets. The model of Word2Vec and BERT have similar performance in news recommendations. However, the BERT model achieves the lowest score in each matrix. 

In our experiment, all wording-level embeddings adopt pre-trained models, such as Word2Vec, GloVe, and BERT. It can be believed that each word has its inherent pre-trained representations, which are nothing about its semantic meanings. Under this exact scenario, the performance of the GloVe and Word2Vec is better than the BERT wording representations in our case. Because our LRM is a pre-trained non-contextual embedding module, a specific contextual embedding-based method, i.e., BERT, has no advantage. Furthermore, models like Word2Vec and GloVe perform better for sentences with many unregistered words \cite{gan2022semGloVe}. Therefore, the BERT method is not suitable for our experiment.

\begin{figure}[t]
\centering
\includegraphics[width=\textwidth]{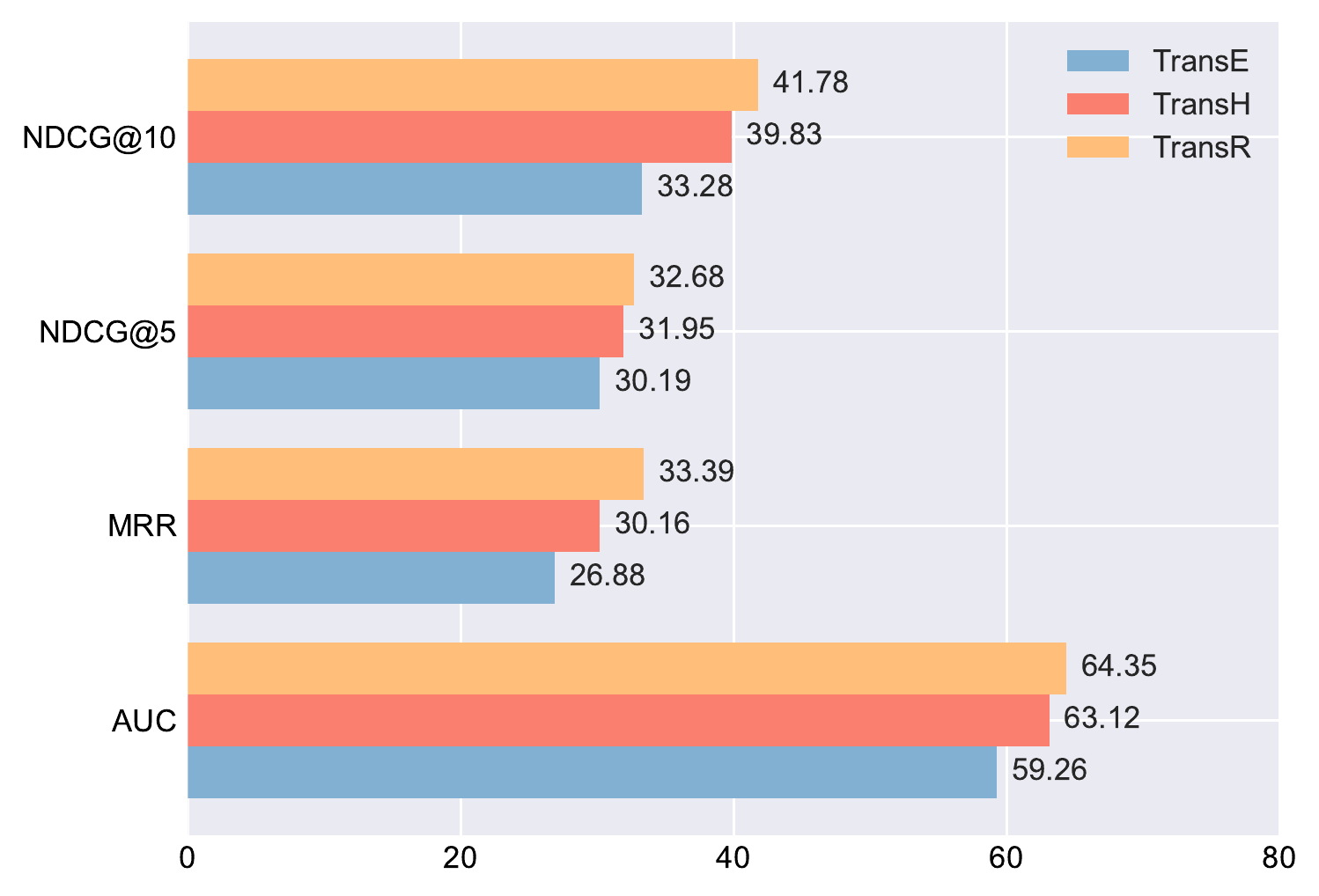}
\caption{Ablation study on different embedding models.}
\label{fig:embed}
\end{figure}
The performance of embedding is crucial for recommendation systems. To evaluate the effectiveness of different embedding methods in our DOR system, we test three different representation models (TransE, TransH, and TransR) to determine the basis of our internal observation mechanism. As shown in Figure \ref{fig:embed}, it is evident that the recommendation performance is most stable when TransR representations (yellow bar) are adopted with the same parameters. All evaluation matrices achieve the highest values when using TransR representation as the input data for our internal observation mechanism. It can be inferred that the translational model, which can handle $1-N$ and $N-N$ relations, is more suitable for our system. In contrast, the TransE model has significant limitations in our case, achieving an AUC of only 59.26.

 \subsection{Parameter Analysis}

    In this section, we investigate the impact of different parameters on our proposed DOR model, including the epoch size and the number of neighbourhoods.
    
\begin{figure}
\centering
\subfigure[AUC score for different epoch sizes]{\includegraphics[width=0.49\textwidth]{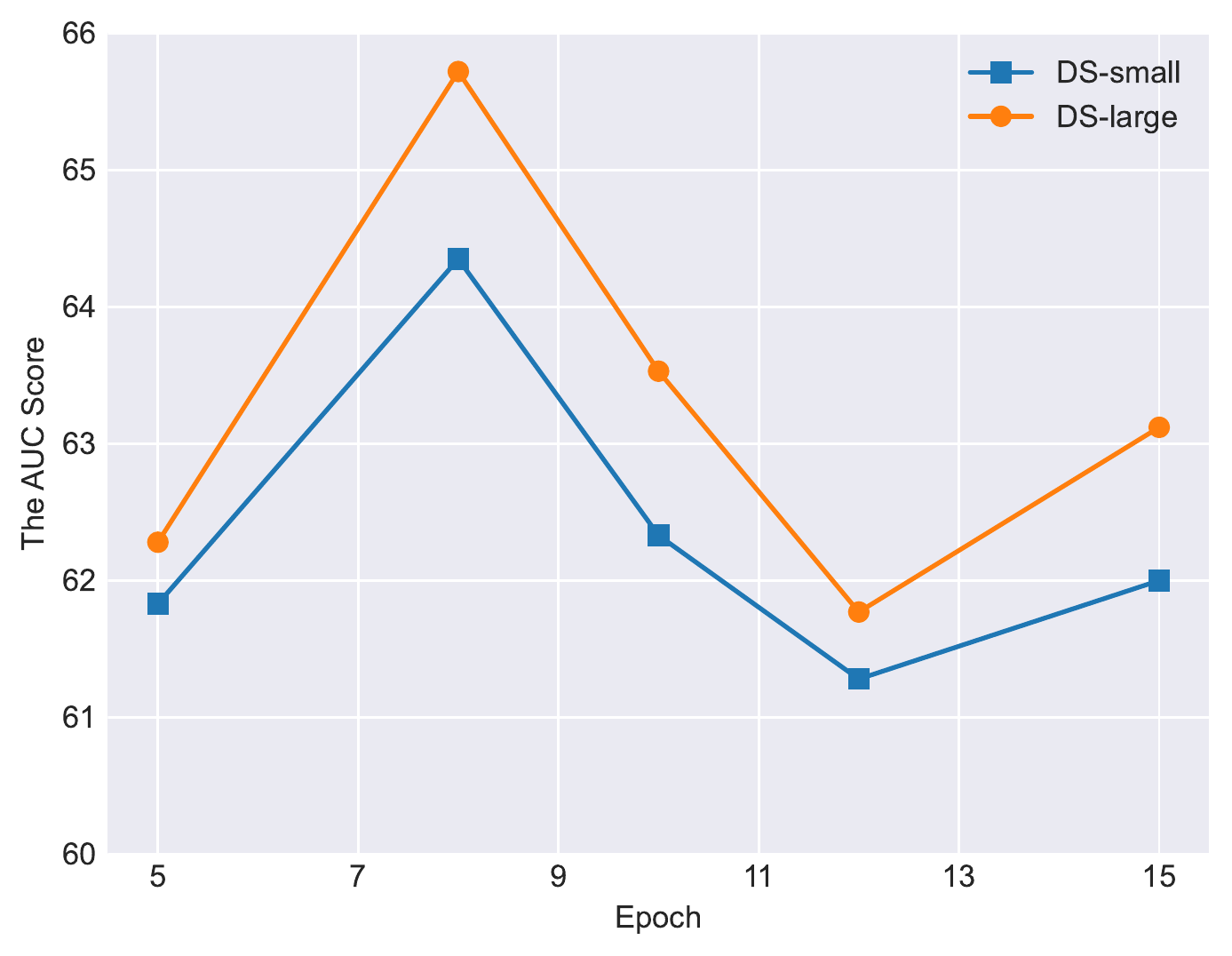}\label{fig:auc_epoch}}
\subfigure[AUC scores for different neighbour numbers]{\includegraphics[width=0.49\textwidth]{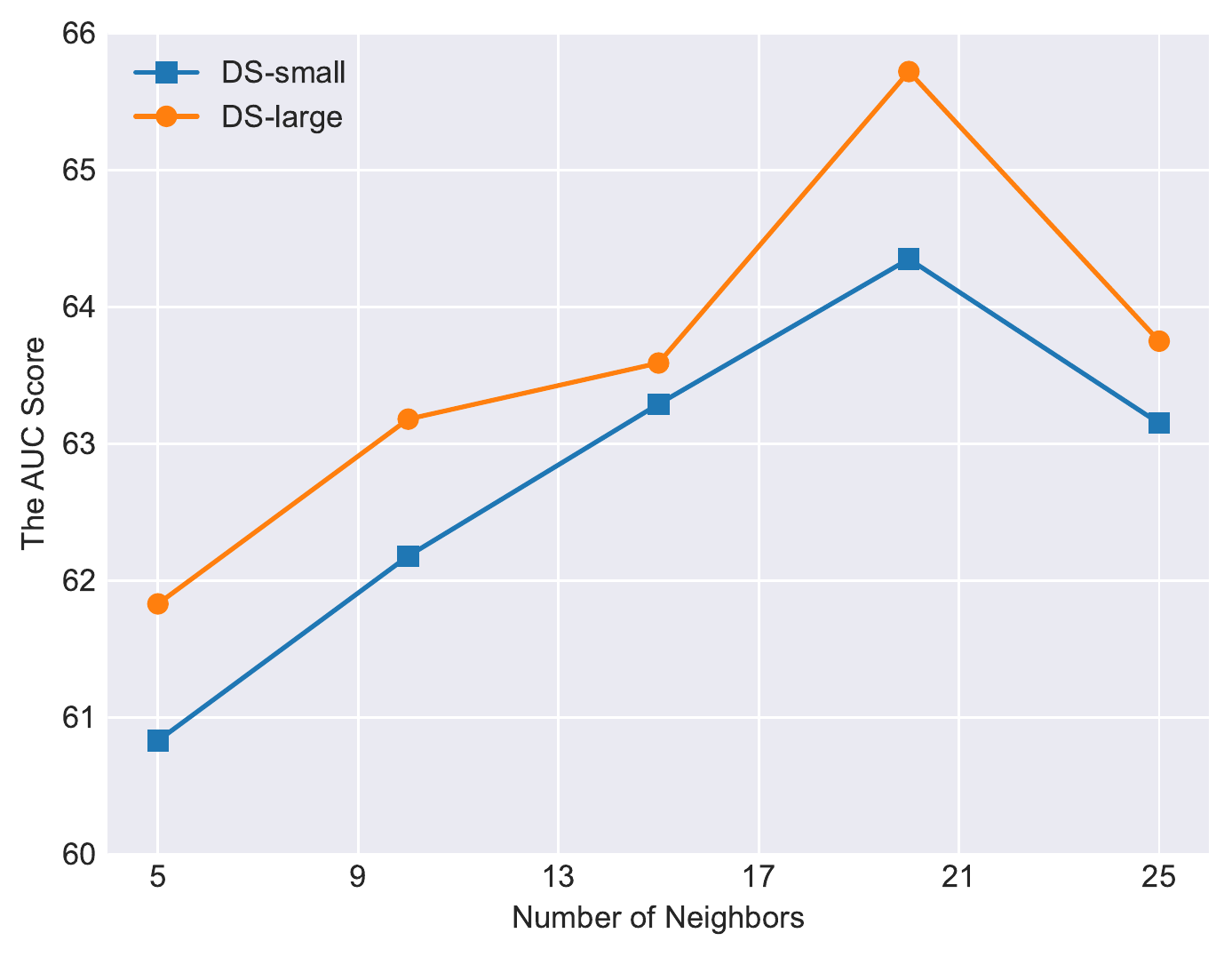}\label{fig:auc_nei}}
\caption{Parameter analysis on different epoch sizes and neighbour numbers}
\label{fig:parameter}
\end{figure}

    Our experiments show that the performance of our model is consistently better with the larger datasets, as shown in Figure \ref{fig:auc_epoch}. We test our model with five different epoch sizes, which are 5, 8, 10, 12, and 15, respectively. The graph demonstrates that the performance starts at an epoch of 5, then increases until 8 epochs, reaching its peak when the model is trained by 8 epochs. However, the performance of our model decreases from 8 epochs to 12 epochs and reaches the worst performance at 12 epochs.
    
    To evaluate the effect of various numbers of neighbourhoods, we provide five different neighbour sizes to our proposed model. As shown in Figure \ref{fig:auc_nei}, the model is most stable when each entity is represented by a maximum of 20 neighbourhoods and their relations. The model's performance keeps a steady increase from 5 to 20 neighbourhoods. However, a decrease appears after 20. Therefore, in our case, we set the number of neighbourhoods to 20 to balance the trade-off between performance and computational complexity.

In summary, our experiments demonstrate that the DOR model holds a clear advantage in news recommendations when utilising the dual-observation mechanism. Our results confirm that it is essential to fully observe the features of news items. Additionally, our experiments demonstrate the importance of considering the mutual influence between the user's belief network and the news in the recommendation process.

\section{Conclusion and Future Work}

    This paper presents a novel approach for news recommendation, called DOR, which combines news representations and user belief networks. DOR comprises two modules: the internal-observation module and the external observation module. The internal observation module extracts news representations from both high- and low-order levels, addressing the limitation of data sparsity and enhancing the semantic meaning of each word in the news. The external observation module also considers the importance of users' belief networks, where the news representation is based on their knowledge and interests. Through extensive experiments and comparisons with various baselines using real-world datasets, the proposed DOR model demonstrates promising stability and accuracy.

    In future research, we aim to further investigate the significance of observation mechanisms in the news recommendation field. We will delve deeper into news representation from high- and low-order interactions and explore triple news representations to represent users' reading behaviours. Additionally, we will examine other news features, such as images and categories, to enhance the performance of the recommendation system.

    \bmhead{Acknowledgments}

    The authors would like to acknowledge the financial support from Callaghan Innovation (CSITR1901, 2021), New Zealand, without which this research would not have been possible. We are grateful for their contributions to the advancement of science and technology in New Zealand. The authors would also like to thank CAITO.ai for their invaluable partnership and their contributions to the project.

	\bibliographystyle{splncs04}
	\bibliography{bibFile}

\end{document}